\documentclass{article}
\usepackage[utf8]{inputenc}
\usepackage{listings}
\usepackage{url}
\usepackage{geometry}
\usepackage{color}
\usepackage{hyperref}
\usepackage{easy-todo}
\usepackage{graphicx}
\usepackage[dvipsnames]{xcolor}
\usepackage{mathtools}
\usepackage{longtable}

\begin{document}
\vspace*{0.2in}

\begin{flushleft}
{\Large
\textbf\newline{Biologically-inspired Salience Affected Artificial Neural Network}
}
\newline

Leendert A Remmelzwaal \textsuperscript{1*},
Jonathan Tapson \textsuperscript{3},
George F R Ellis \textsuperscript{2},
Amit K Mishra \textsuperscript{1} \\

\bigskip
\textbf{1} Department of Electrical Engineering, University of Cape Town, Rondebosch, Cape Town,
South Africa 7700.
\\
\textbf{2} Department of Mathematics and Applied Mathematics, University of Cape Town, Rondebosch, Cape Town, South Africa 7700.
\\
\textbf{3} MARCS Institute for Brain, Behaviour and Development, Western Sydney University, Sydney, Australia.
\\
\bigskip

* Corresponding author: leenremm@gmail.com

\end{flushleft}


\section*{Abstract}

In this paper we introduce a novel Salience Affected Artificial Neural Network (SANN) that models the way neuromodulators such as dopamine and noradrenaline affect neural dynamics in the human brain by being distributed diffusely through neocortical regions, \textcolor{black}{allowing both salience signals to modulate cognition immediately, and one time learning to take place} through strengthening entire patterns of activation at one go. We present a model that is capable \textcolor{black}{of \textit{one-time salience tagging} in a neural network trained to classify objects}, and returns a \textcolor{black}{\textit{salience response}} during classification \textcolor{black}{(inference)}. We explore the effects of salience on learning via its effect on the activation functions of each node, as well as on the strength of weights \textcolor{black}{between nodes} in the network. We demonstrate that salience tagging can improve \textcolor{black}{classification confidence for both the individual image as well as the class of images it belongs to}. \textcolor{black}{We also show that the computation impact of producing a salience response is minimal}. This research serves as a proof of concept, and could be the first step towards introducing salience tagging into Deep Learning Networks and robotics.


\section*{Glossary}

\noindent{AI = artificial intelligence} \\
\noindent{ANN = artificial neural network} \\
\noindent{BBD = brain-based device} \\
\noindent{GCNet = global context network \cite{Cao2019Gcnet}} \\
\noindent{LTD = long-term depression} \\
\noindent{LTP = long-term potentiation} \\
\noindent{MSE = mean squared error} \\
\noindent{NN = neural network} \\
\noindent{PLU = Piecewise Linear Unit (activation function)} \\
\noindent{PRP = plasticity-related proteins \cite{Redondo2011taggingcapture}} \\
\noindent{ReLU = rectified linear unit} \\
\noindent{RNN = recurrent neural network} \\
\noindent{SANN = salience-affected neural network} \\
\noindent{SENet = squeeze-and-excitation network \cite{Hu2018squeeze}} \\


\newpage
\section{Introduction}

This paper introduces a new kind of artificial neural network (ANN) architecture, namely a \textit{Salience Affected Artificial Neural Network (SANN)}. The SANN models the effect of neuromodulators in the cortex \cite{ellis2005neural} \cite{panksepp2004affective} \cite{Panksepp1982emotions} \cite{Panksepp2003emotions} \cite{Panksepp2005emotions}: an important feature of the human brain, based on the well established fact that emotions play a key role in brain function, see for example the writings by Antonio Damasio such as \textit{Descarte’s Error} \cite{damasio1995descartes} and \textit{The Feeling of What Happens} \cite{damasio1999feeling}. This SANN architecture gives powerful additional functionality to SANNs that are not possessed by other ANNs\textcolor{black}{, namely the addition of one-time salience tagging, salience response during inference and improved classification confidence (see Section \ref{sec:results})}. It could be a key component of approaches to artificial intelligence that involves designing machines with feeling analogues \cite{damasio_machines}.

\subsection{\textcolor{black}{The basic underlying assumptions}}

There are two basic assumptions underlying this paper. Firstly, evolution has fine-tuned human brain structure over millions of years to give astonishing intellectual capacity. It must be possible for designers of ANNs to learn possible highly effective neural network architectures from studying brain structure  \cite{edelman2007learning}. This has of course happened in terms of the very existence of ANNs, which are based in modeling the structure of cortical columns. It has not so far happened as regards the structure and function of the ascending systems considered here (an exception is Edelman himself who modelled them, but he left out key aspects as we discuss below). However they have been hardwired into the human brain by evolutionary processes precisely because they perform key functions that have greatly enhanced survival prospects. This architecture should therefore have the capacity to significantly increase performance of any kind of ANN, and so has the potential to play an important role in robotics or AI. \\

Secondly, while the brain is immensely complex and therefore requires study at all scales of detail in order that we fully understand it, nevertheless it can be claimed that there are basic principles that characterise its overall structure and function, that can be very usefully developed in simple models such as presented here. These models can provide an in-principle proof that the concept works, and so it may be worth incorporating this structure in much more complex models such as massive deep learning or reinforcement learning networks. The testing we do on the simple models presented here suggests that may indeed be the case.

\subsection{\textcolor{black}{Key features of the SANN}}

The key feature modelled here is the way neuromodulators such as dopamine and noradrenaline are distributed diffusely through neocortical regions by `ascending systems’ originating in nuclei in pre-cortical areas (see Fig \ref{fig:Fig003}). These connections contrast with the highly specific synaptic connections between neurons in neocortical columns, which of course also occur and are modelled in standard ANNs. The ascending systems are not connections to specific neocortical neurons: rather they spread neuromodulators to all synapses in specific cortical regions. \\

\textcolor{black}{Neuromodulators released in the cortex by the arousal system} affect an entire pattern of synaptic connections that are active in that region at that time by altering their weights in proportion to the product of the synaptic level of activity and the strength of the neuromodulator released. \textcolor{black}{In addition to having a lasting impact on a learned pattern, salience also impacts the cortex at the time the neuromodulator is released (during salience tagging) and the effects on action and attention are immediate.} This is the extremely powerful mechanism, described in the book \textit{Neural Darwinism} by Gerald Edelman \cite{edelman1987neural}, which strengthens an entire pattern of cortical activation at one go. That is what enables one-time learning to occur. The link to emotions, and so affect, is because these ascending systems are also the physiological basis of the genetically determined \textcolor{black}{primary emotional systems \cite{panksepp2004affective} \cite{Panksepp1982emotions} \cite{Panksepp2003emotions} \cite{Panksepp2005emotions}. Next, we discuss four key features of salience: (1) salience tagging of memories, (2) the impact of salience tagging on classification confidence, (3) salience retrieval during classification, and (4) the impact of salience on attention, decision making and the desire to act (e.g. fight or flight).}

\subsubsection{Salience tagging}

\begin{figure}[ht!]
    \centering
    \includegraphics[scale=0.6]{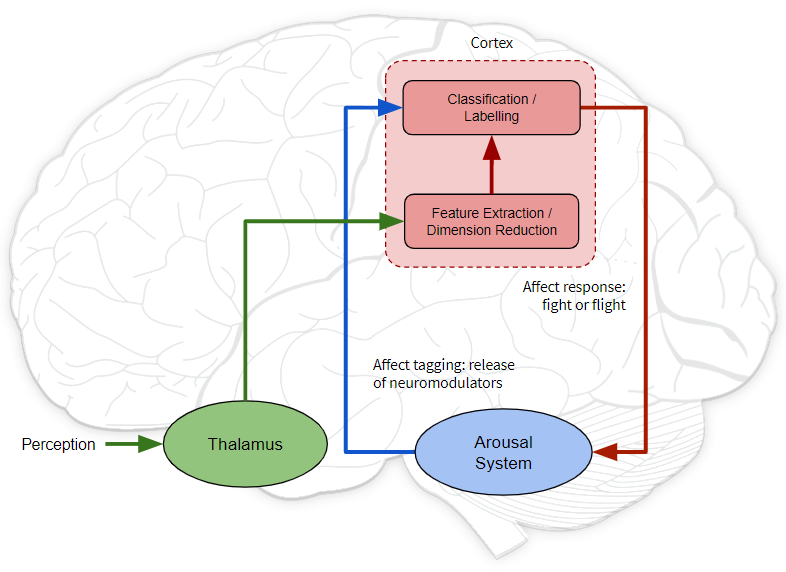}
    \caption{\textit{The structures involved in emotional tagging: The Thalamus receives incoming sensory input, which is passed to low level processing in the cortex (feature extraction). Thereafter the information is passed to higher level classification/labelling regions in the cortex. The arousal system controls the release of neuromodulators, and is responsible for affect tagging as well as processing affect response.}}
    \label{fig:brain}
    \end{figure}

Firstly, this SANN architecture allows for \textit{one-time salience tagging of memories} to take place, by affecting entire patterns of activation in one go. A memory trace of a specific event is formed and stored together with a salience tag associated with the event, which could be positive or negative \cite{kim2007neural} \cite{touboul2017affects}, depending on its nature. \textcolor{black}{The key structures involved in this process in the cortex are the thalamus, the neocortex and the arousal system (see Fig \ref{fig:brain})}. \\

\begin{figure}[!h]
     \centering
     \includegraphics[width=0.7\linewidth]{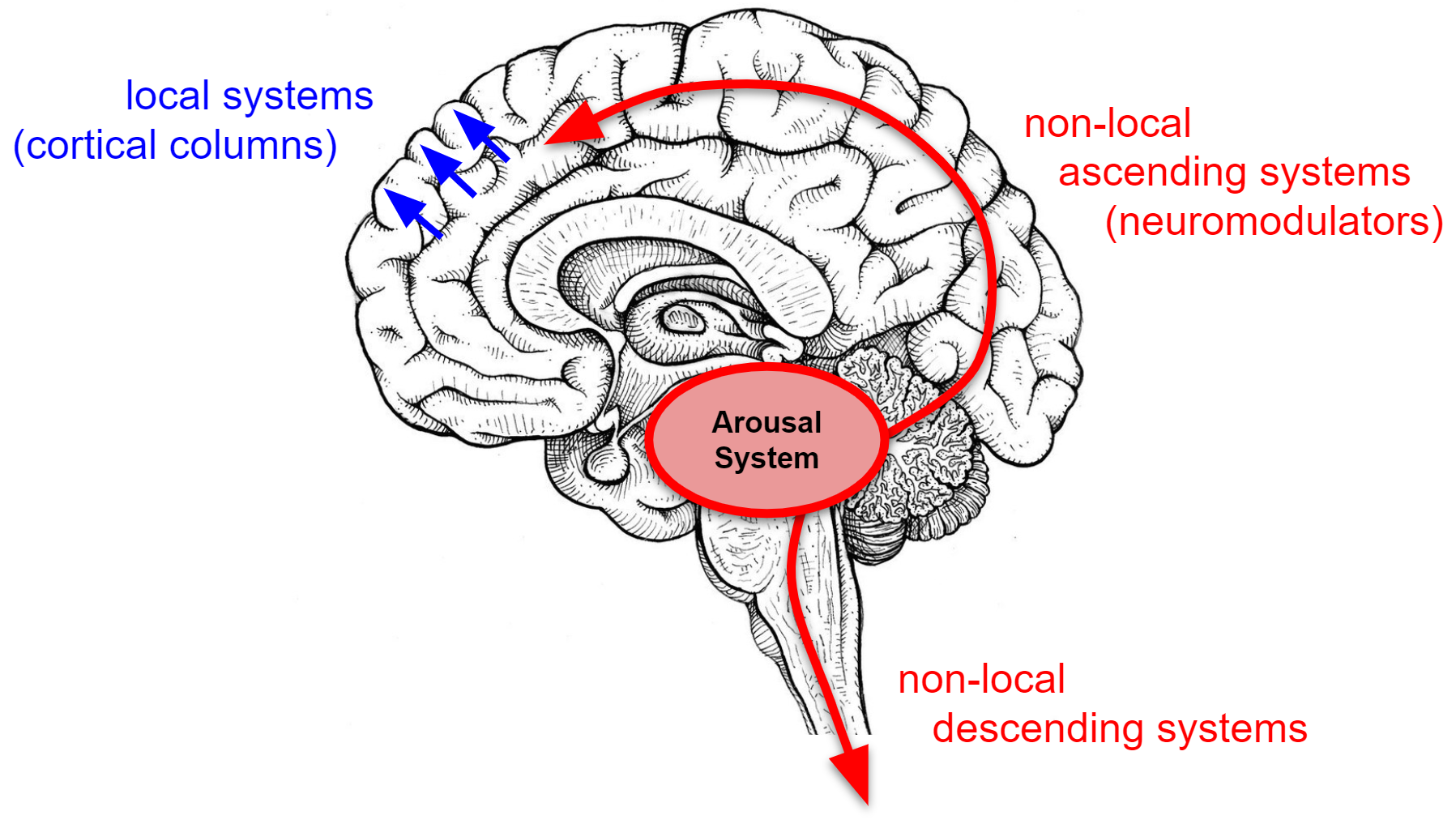}
     \caption{\textit{There are two very different kinds of circuits at work in the cortex that we consider here. Firstly there are  local systems, such as signals sent between layers in cortical columns. But there are also non-local systems ascending from the arousal system (e.g. the limbic system). The arousal system sends neuromodulators (e.g. dopamine and noradrenaline) into the cortex by means of diffuse projections. These neuromodulators simultaneously affect entire patterns of activated neurons at the time the neuromodulators are delivered, acting on the memory of the object currently being observed.}}
     \label{fig:Fig003}
     \end{figure}

One of the ways in which memories are tagged with salience in the cortex is via changes in synaptic strength due to neuromodulators \cite{Nadim2014synapticstrength} \cite{Seol2007synapticstrength} \cite{Bucher2013synapticstrength} that are spread diffusely from the excitatory system (based in pre-cortical areas) to the neocortex via ascending system \cite{edelman1987neural} \cite{edelman2004wider} (see Fig \ref{fig:Fig003}). The one-time learning with this SANN architecture contrasts crucially with the thousands if not millions of repetitions needed to train a neural network to correctly classify objects via back-propagation and similar methods of adjusting neural network weights, as in usual ANNs. \\

Neuromodulators have the effect of tagging existing memories with salience after only a single iteration of salience training. It is important to highlight that neuromodulators are not responsible for creating new neural connections on their own; they can only strengthen existing pathways with salience; for example by promoting LTP or LTD \cite{Frey1997ltp} \cite{ODonnell2012norepinephrine} \cite{Durstewitz2000models}. \textcolor{black}{Assuming that a base layer of classification training has been completed prior, salience training has a positive impact on both the individual and the class, and is therefore an alternative to additional iterations of back propagation. In this paper we explore the effects of salience on a classification neural network, but this can be extended to model the changes in synaptic strength related to long-term potentiation (LTP) which is the primary cellular model of memory in mammals \cite{Frey1997ltp}. Other key memory-related mechanisms which we do not model in this paper include: distributed associative storage, plasticity-related proteins (PRP), and the capture of these proteins by tagged synapses \cite{Redondo2011taggingcapture}. These would be suggested areas of future research.} \\

It is important to note that the one-time learning considered here is not the same as `one-shot’ learning \cite{fei2006one} \cite{vinyals2016matching} which relates to learning categories of entities or events. In contrast `one-time' learning in this paper refers to the tagging of specific individual memory with salience, so that a specific memory is associated with an emotional tag.

\subsubsection{Salience improves \textcolor{black}{classification confidence}}

Thirdly, the SANN architecture models the impact that diffuse neuromodulators in the cortex have on neurons; affecting the activation functions of neurons, as well strengthening the synaptic weights. This study demonstrates how a salience signal can improve the classification \textcolor{black}{confidence of the salience-tagged image, for images in the same class, as well as the network as a whole}.

\subsubsection{Salience retrieval}

Secondly, whenever a salience-tagged memory is retrieved, the salience tag is retrieved along with the memory, alerting the cortex to its significance and priming it to again act appropriately in response. The SANN architecture allows emotional tags to be attached to sensory inputs, signifying their importance and how to react to them, and recovers these emotional signals when the image is identified at another occasion. \textcolor{black}{For example if one sees} a man approach threateningly with a knife in his hand, one experiences the corresponding perceptions (sight, sound, etc) together with a salience (or significance) signal which focuses attention on that event and associated features. That salience signal, experienced as emotional feelings, is an indication that these are important issues for welfare or even survival that must be dealt with right now. The mind puts other issues aside, and decides how to handle this event in a safe way. This is the immediate effect on attention, and so changes cognitive processes by causing attention to have a specific focus (thinking about a knife rather than walking to the bus stop). \\

\textcolor{black}{In addition to the subject of a scene, contextual cues also impact the salience response of an object. For example, a knife perceived by an attacker has a negative salience, however, a knife on a cutting board may evoke a positive salience response if you enjoy cooking. We believe that it would be possible that a SANN could use such crucial contextual cues to disambiguate salience signals and hence develop holistic scene understanding. In this paper we focus only on the subject of a scene, but extending this research to contextual cues would serve as a valuable extension of this research.} \\

\textcolor{black}{Salience retrieval may have an application in solving the nearest neighbour problem \cite{Andoni2006near} (given a collection of \textit{n} points, build a data structure which, given any query point, reports the data point that is closest to the query) or enhancing the locality-sensitive hash function \cite{Dasgupta2017hash}. In both cases a quick response returning an approximate set of nearest-neighbors is favoured over accuracy, and the SANN has the benefit of speed and performance (see Section \ref{sec:results}). We suggest this as a future extension of the proof-of-concept we present in this paper.}

\subsubsection{Desire to act}

\textcolor{black}{The salience response retrieved during classification} thus gives key guidance as to future actions; it drives living organisms to act in either an attractive (in the case of pleasure) or repulsive way (in the case of fear). \textcolor{black}{Lövheim describes the role of noradrenaline as ``coupled to the fight or flight response and to stress and anxiety, and appears to represent an axis of activation, vigilance and attention" \cite{Lovheim2012cube}. The fight or flight response has been implemented in other well known cognitive architectures \cite{vallverdu2016neucogar} \cite{remmelzwaal2020brainCSR} but in this paper we simplify this to a single value called ``desire to act" (see Section \ref{sssec:drivetoact}). In future research this ``desire to act" value could connect with the fight or flight response in a selected cognitive architecture.}

\subsection{Scope of this paper}

\textcolor{black}{In this paper we investigate all three aspects noted above.} After describing this architecture, the features noted will be demonstrated by a systematic exploration of examples, using the publicly available MNIST \textcolor{black}{dataset of handwritten letters \cite{yann1998mnist} and the animal silhouette dataset \cite{remmelzwaal2020brainCSR}.}

\subsection{Limitations}

\textcolor{black}{This paper is intended to present a proof-of-concept of the SANN model. We do not model what triggers the salience signal,} \textcolor{black}{and we only begin to model the impact salience could have on cognition and attention.} These undoubtedly need to be incorporated in a more complete model of the brain processes of interest, but in order for our investigation to be focused and manageable we concentrate on the effects mentioned above. The relation to perception and attention will be the subject of future papers. We also do not attempt to model spiking neural networks; that again can be the subject of future investigation.

\subsection{Structure of this paper}

\textcolor{black}{In Section \ref{sec:related_work} we discuss related work, and in Section \ref{sec:method} we describe the SANN model architecture.} In Section \ref{sec:experimental_design} we describe the experimental design, and in Section \ref{sec:results} we present the results of the simulations we ran and the observations made. We then draw conclusions in Section \ref{sec:discussion}, and make suggestions for future work in Section \ref{sec:future_work}.


\newpage{}
\section{Related work} \label{sec:related_work}

There have been a number of attempts to model non-local effects in neural networks and robotics. In this section we review related works. \\

\textcolor{black}{In 1998, Husbands presented GasNets to model the presence of the NO gas in the environment surrounding the neurons, which is capable of non-locally modulating the behaviour of other nodes \cite{husbands1998evolving}. Like SANNs, this form of modulation allows a kind of plasticity in the network in which the intrinsic properties of nodes are changing as the network operates. To the best of our knowledge, GasNets have not been modified to train and test an ANN with specific salience, nor have they been used to demonstrate one-time learning in ANNs.} \\

\textcolor{black}{In 2005, Gerald Edelman created a range of brain-based devices (BBDs) including Darwin VII and Darwin X \cite{edelman2007learning} \cite{krichmar2005brain}. Edelman's BBD models required many iterations of training for the salience to be embedded in the model and associated with sensory input patterns. Edelman's model was also not able to demonstrate one-time learning on an previously trained dataset, which is what we demonstrate in our SANN.} \\

\textcolor{black}{In 2009, Khashman presented an emotional neural network (EmNN) which he applied to the application of blood cell identification \cite{Khashman2009blood}, and later to the application of credit risk scores in 2011 \cite{Khashman2011credit}. Khashman introduced emotional neurons to a neural network as a separate set of neurons in the model, instead of embedding emotion in the existing neurons as we have done in the SANN model. This means that the EmNN model requires the emotional neurons to be updated during standard classification training alongside the regular neurons, not as a single-shot learning as we do in our model. The SANN model we present in this paper follows a bio-realistic implementation, with associated computational efficiencies and benefits.} \\

\textcolor{black}{In 2013, Thenius presented another model of emotions in an artificial neural network called the EMANN \cite{Thenius2013EMANN}, where he models the effect of hormones at affecting the weights, net-functions, and out-functions. This model received minimal experimental testing; a network with 4 nodes that solves an simple XOR problem. In our work we implement emotional tagging on more complex applications. Thenius only applied a single dimension of hormone, while we investigate a complementary pair of neuromodulators: dopanine and norepinephrine. In the EMANN model, the hormone affected the hill function as a fixed offset, whereas we explore multiple effects on the activation function. Lastly, Thenius applied the effect of the hormone during initial training of the EMANN model. By comparison, we explore the impact of one-time learning after standard classification training, modelling the effect of releasing a neuromodulator in the cortex as an adult.} \\

\textcolor{black}{A similar concept to `salience' is the `attention mechanism'; a mechamism used commonly with Neural machine translation (NMT) \cite{kalchbrenner2013recurrent}. Attention mechanisms applied to translation activities can improve translations by selectively focusing on parts of the source sentence during translation \cite{luong2015effective} \cite{bahdanau2014neural}. In 2017, Wang introduced a trunk-and-mask attention mechanism using an hourglass module to generate attention-aware features \cite{Wang2017trunk}, and in 2018 Hu presented the Squeeze-and-Excitation model performs \textit{feature re-calibration}, whereby it learns to use ``global information to selectively emphasise informative features and suppress less useful ones" \cite{Hu2018squeeze}. In 2018, Wang presented non-local model to improve long-range dependencies \cite{Wang2018nonlocal}; a generalization of the classical non-local mean operation \cite{Buades2005nonlocal}, and aims to extract the global understanding of a visual scene. Wang's non-local model was initially applied to tasks of video classification, object detection and segmentation, and pose estimation. Cao's Global Context Network (GCNet) combines the Non-local Networks (NLNets) and Squeeze-and-Excitation Networks (SENet) \cite{Hu2018squeeze} to create a global (query-independent) attention map \cite{Cao2019Gcnet}. In each of these cases, the attention mechanism is achieved by swapping out specific layers in the neural network with alternative neural network layers \cite{kim2017structured} \cite{vaswani2017attention} \cite{Hu2018squeeze} \cite{Wang2017trunk}. By comparison, the SANN does not swap out or scale hidden representation layers dynamically and instead implements an additional salience variable on top of a standard ANN, without changing the dimensions of the hidden representations.} \\

\textcolor{black}{In addition to the attention mechanisms mentioned above, computational-neuroscience models of visual saliency have been developed to produce a single saliency map \cite{Itti2001visualattention}, highlighting which specific features in an image are considered the most salient \cite{Linsley2018learning}. Saliency maps have been used in image processing to improve accuracy and speed in tasks such as object detection or pose estimation. Our work differs from the field of saliency maps, as we apply a salience to an entire image and model one-time salience tagging and salience response.}  \\

In contrast to other related work, in this paper we demonstrate how the `salience' signal from the arousal system (which Edelman called the `value' system) can be modelled as a \textcolor{black}{single} additional dimension to a \textcolor{black}{standard artificial neural network (ANN)}. \textcolor{black}{This single dimension allows us to model the effects of dopamine (pleasure) with a positive salience value, and norepinephrine (fear) with a negative value.} We demonstrate how a salience signal can affect a specific neural activation patterns (e.g. a memories), tagging it with a salience signal (e.g. an emotional response) during one-time learning. It is important to note that this research is not the same as some ``emotional robots'' that are constructed so as to physically simulate emotional expressions. We rather \textcolor{black}{model the release of diffuse neuromodulators in the cortex, and the impact this has on nodes and weights in an ANN.} One could additionally add the aspect of emotional expression simulation if desired, but this falls outside of the scope of this research.


\newpage
\section{Method} \label{sec:method}

\subsection{The concept}

\textcolor{black}{The SANN and supporting architecture are modelled after the non-local systems ascending from the arousal system (see Fig \ref{fig:Fig003}). The specific cortical structures modelled are the thalamus, the arousal system and the neo-cortex \cite{remmelzwaal2020brainCSR} as shown in Fig \ref{fig:brain}.}

    \begin{figure}[ht!]
    \centering
    \includegraphics[scale=0.7]{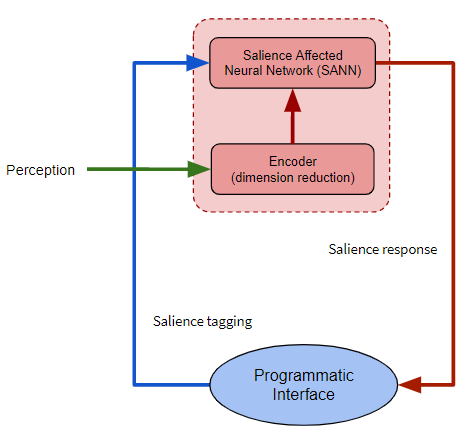}
    \caption{\textit{\textcolor{black}{Conceptual overview of proposed model: Input images are passed to an encoder first, before being passed to a SANN model. The salience tagging and salience response are available via a \textcolor{black}{programmatic interface (for the purposes of running simulations)}.}}}
    \label{fig:model}
    \end{figure}

\textcolor{black}{The implementation of this architecture is described in Fig \ref{fig:model}. We first process the incoming sensory input (visual) using an Encoder for dimensional reduction, before passing the Encoded representation of the input to a SANN for classification and salience tagging. Lastly, the salience tagging and monitoring of the salience response are both managed with a manual interface. We chose to split the CNN and SANN into 2 separate models so that we could evaluate the effect of salience on an ANN. Future research includes applying salience to other ANN models such as CNNs or RNNs.} \\

\noindent{}\textcolor{black}{In the following section we discuss the mathematical representation of the SANN.}


\subsection{Mathematical representation} \label{sec:modelling_salience}

\begin{figure}[ht!]
    \centering
    \includegraphics[scale=0.35]{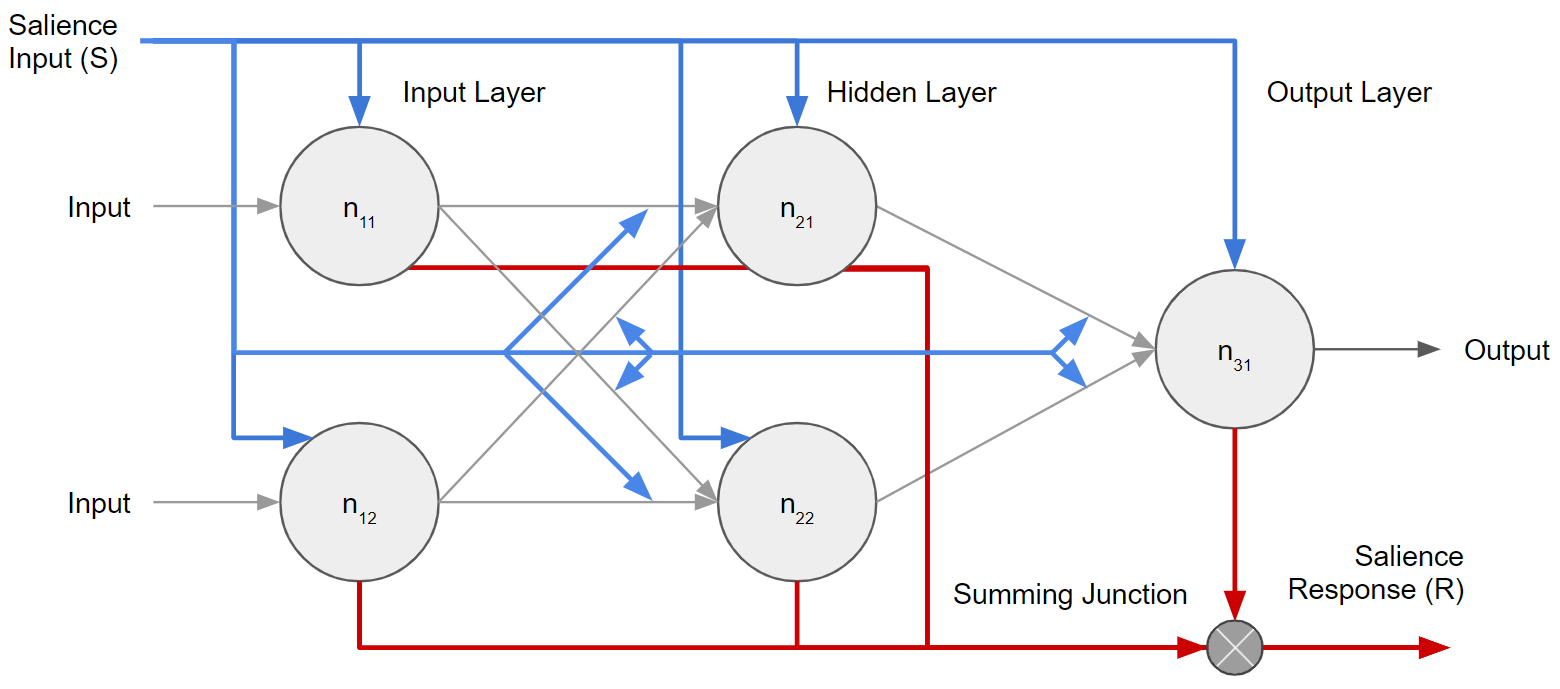}
    \caption{\textit{Schematic illustrating how the salience signal $S$ affects nodes and weights  in the SANN during one-time salience training \textcolor{black}{(blue)}, and how the SANN produces a salience response $R$ during classification \textcolor{black}{(red)}.}}
    \label{fig:Fig002}
    \end{figure}

\textcolor{black}{The salience signal affects both the nodes and the weights during one-time salience training, as illustrated in Fig \ref{fig:Fig002}. In this section we describe the effect of salience on the nodes and weights of the network using formal mathematical notation.} \\

\subsubsection{Salience value}

\textcolor{black}{Each node in the network is assigned a salience value $S_{i}$ in range $[-1,1]$ initially set to $0$. During one-time salience training, this salience value $S_{i}$ of the $i$th node is updated as follows:}

    \begin{equation}\label{eq000}
        S_{i}(N) = S_{i} + (1 - S_{i}) \alpha_{i} N_{i}
    \end{equation}
    \vspace{1mm}

\noindent{}\textcolor{black}{where $N$ represents the level of neuromodulator released, and $\alpha_i$ represents the activation of the node a the time of one-time salience training.} \\

\subsubsection{Positive and negative salience}

\textcolor{black}{While the SANN only has a single salience dimension, we leverage the sign of the salience to model the effects of two separate neuromodulators: dopamine (pleasure) and norepinephrine (fear). We model dopamine as a salience signal with a positive salience value, and norepinephrine with a negative salience value.}

\subsubsection{Strengthening weights}

\noindent{}\textcolor{black}{The weights $W_{i,j}$ of $j$th input synapse of cell $i$ are updated during one-time salience training as follows:}

    \begin{equation}\label{eq100}
        W_{i,j}(S) = W_{i,j} \times (1 + |S_{i} \alpha_{i} \theta |)
    \end{equation}
    \vspace{1mm}

\noindent{}\textcolor{black}{where $S_{i}$ represents the salience value of node $i$, $\alpha_i$ represents the activation of the node $i$ a the time of one-time salience training, and $\theta$ is a constant.} \\

\subsubsection{Impact on activation functions} \label{sssec:sigmoidal}

\textcolor{black}{In addition to impacting the strength of weights, one-time salience training also impacts the activation functions. In this paper we chose to use the sigmoidal activation function because (1) sigmoidal activation functions are more biologically realistic: most biological systems saturate at some level of stimulation (where activation functions like ReLU do not), and (2) sigmoidal functions allow for bipolar activations, whereas functions like ReLU are effectively monopolar and hence not useful for a salience  signal with both polarities of activation. However, this research could be extended to include other activation functions such as Maxout functions \cite{goodfellow2013maxout}, ReLU functions \cite{nair2010rectified} or PLUs \cite{torkamani2019learning}.} \\

\begin{figure}[!h]
     \centering
     \includegraphics[scale=0.5]{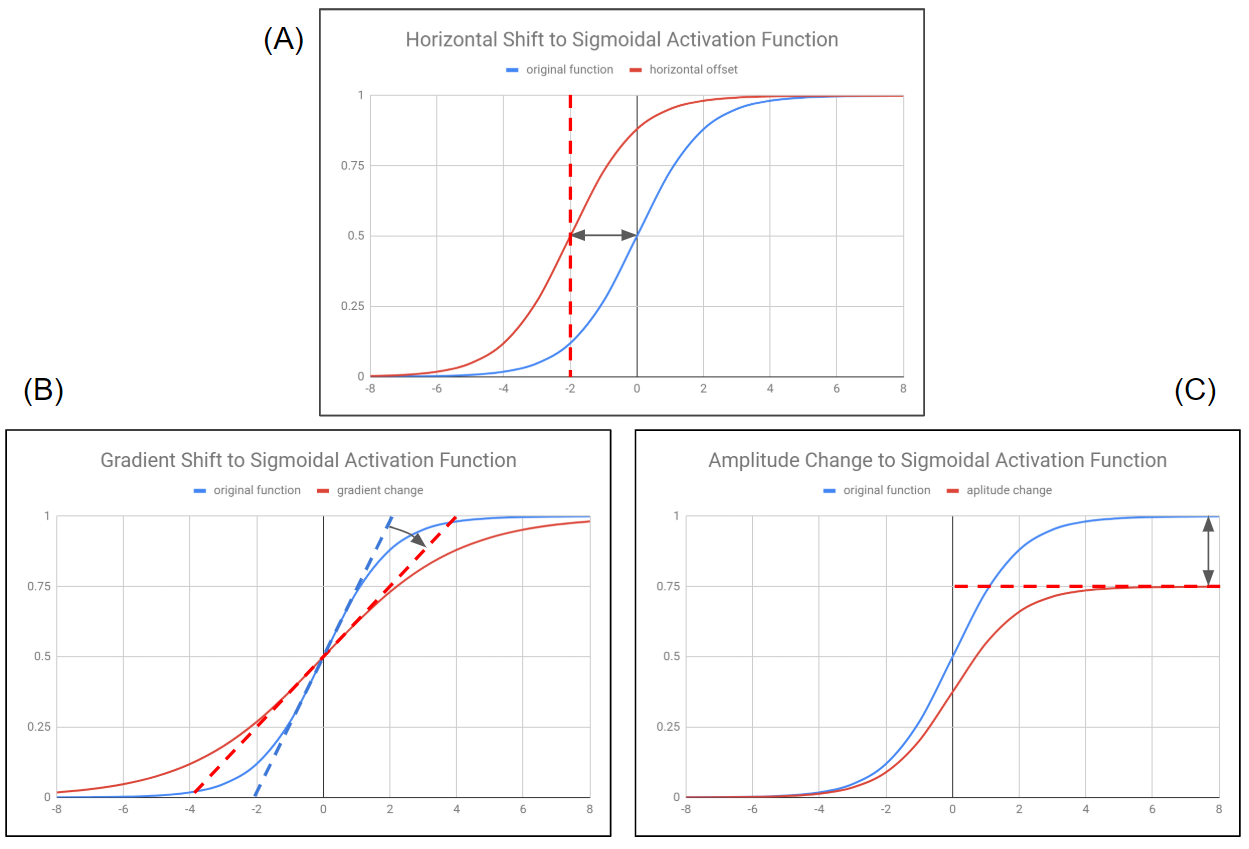}
     \caption{\textit{Three changes to the activation function were explored: \textcolor{black}{(A) Change in the horizontal offset of the activation function, (B) a change in the gradient of the activation function and (C) a change in the amplitude of the activation function: A positive salience signal would increase the amplitude of the activation function, resulting in a higher output from the activation the next time.}}}
     \label{fig:Fig010}
     \end{figure}

\textcolor{black}{We explore three modifications to the activation function, namely 
(1) Horizontal offset, (2) Change in the gradient, and (3) Change in the amplitude. These variations have been visualized in Fig \ref{fig:Fig010}.}

\noindent{}\textcolor{black}{The mathematical representation for a standard sigmoidal activation function is shown in Eq \ref{eq_act}.}

    \begin{equation}\label{eq_act}
        y(x) = \frac{\mathrm{1} }{\mathrm{1} + e^{-x} }
    \end{equation}
    \vspace{2mm}

\noindent{}\textcolor{black}{The horizontal offset of the activation function along the x-axis (Fig \ref{fig:Fig010}A) is by Eq \ref{eq_act_offset_pos}.}

    \begin{equation}\label{eq_act_offset_pos}
        y(x) = \frac{\mathrm{1} }{\mathrm{1} + e^{-(x + S_i)} }
    \end{equation}
    \vspace{2mm}

\noindent{}\textcolor{black}{The gradient change of the slope (Fig \ref{fig:Fig010}B) is described by Eq \ref{eq_act_gradient_b}.}

    \begin{equation}\label{eq_act_gradient_b}
        y(x) = \frac{\mathrm{1} }{\mathrm{1} + e^{-(x) \times \sqrt{0.5 ^ {-S_i}}}}
    \end{equation}
    \vspace{2mm}

\noindent{}\textcolor{black}{The amplitude change of the slope (Fig \ref{fig:Fig010}C) is described by Eq \ref{eq_act_amp}.}

    \begin{equation}\label{eq_act_amp}
        y(x) = \frac{0.5 ^ {-S_i} }{\mathrm{1} + e^{-x}}
    \end{equation}
    \vspace{2mm}

\noindent{}\textcolor{black}{In all three variations of the activation function, $S_i$ can be either a positive or negative value. A positive value models the positive emotion of pleasure associated with dopamine, while a negative value models a emotion fear associated with norepinephrine.} \\

\subsubsection{Salience response}

\noindent{}\textcolor{black}{During classification, the salience response $R$ produced by the SANN is calculated as follows:}

    \begin{equation}\label{eq200}
        R(S) = \sum_{i=1}^{n} S_{i} \times \alpha_{i}
    \end{equation}
    \vspace{2mm}

\noindent{}\textcolor{black}{where $S_{i}$ represents the salience value of node $i$ and $\alpha_i$ represents the activation of the node $i$ a the time of one-time salience training.} \\

\subsubsection{Classification confidence} \label{sssec:confidence}

\noindent{}\textcolor{black}{The SANN was designed as a Sigmoid classifier with activation functions described in Section \ref{sssec:sigmoidal}. We calculated the classification confidence as the associated confidence $\hat{P}$ of the class prediction $\hat{Y}$. This could be extended in future research to a \textit{calibrated confidence} as described by Guo \cite{Guo2017confidence}.}

\subsubsection{Desire to act} \label{sssec:drivetoact}

\textcolor{black}{In the SANN we model the desire to act $D$ as a value proportional to the salience response $R$ produced by the SANN model after classification. A formal mathematical representations for the desire to act $D$ would be:}

\begin{equation}\label{eq300}
    D(R) = \gamma \times R \\
\end{equation}

\begin{equation}\label{eq301}
    D(S) = \gamma \times \sum_{i=1}^{n} S_{i} \times \alpha_{i}
\end{equation}
\vspace{1mm}

\noindent{}\textcolor{black}{where $\gamma$ is a constant, and $R$ is the salience response and $S_{i}$ represents the salience value of node $i$.} \textcolor{black}{What we notice here is that salience affects actions directly at the time of salience tagging, as well as when a salience response is produced during future classification.} \\

\textcolor{black}{The desire to act $D$ value can be linked to an action module to guide actions. The implementation of this falls outside of the scope of this paper, but an example of this already exists; the SANN model was embedded in the CODA cognitive architecture \cite{remmelzwaal2020brainCSR} to manage the tagging and response of affect.}

\subsubsection{Propagation and bias}

\noindent{}\textcolor{black}{The propagation and bias functions are not affected by the salience training.}


\newpage
\section{Experimental design} \label{sec:experimental_design}

\textcolor{black} {In the previous section we have discussed the SANN model conceptually and mathematically. In this section we shall present the experimental process to implement the SANN architecture.}

\subsection{Key merits}

\textcolor{black} {We shall endeavour to quantify and validate some of the major benefits of SANN compared to the standard ANN back-propagation learning algorithm. The following are some of the major merits of the SANN:}

\begin{enumerate}
    \item \textcolor{black} {The SANN is capable of one-time salience training.}
    \item \textcolor{black} {One-time salience training affecting the weights and activation functions should result in an increase in classification confidence for the individual object tagged with salience, as well as the entire class the tagged object belongs to.}
    \item \textcolor{black} {The intensity of salience signal during salience tagging to be positively correlated with increased classification confidence.}
    \item \textcolor{black} {We expect the salience response calculation (during inference) to have a relatively minor impact on the performance (i.e. less than a 10\% increase).}
\end{enumerate}

\subsection{Methodology}

\noindent \textcolor{black} {In the Table \ref{tab:method}, we present the methodology through which we shall validate the merits listed above.}

    \begin{table}[ht]
    \renewcommand{\arraystretch}{1.5}
    \begin{center}
    \begin{tabular}{ |p{2cm}|p{12cm}|}
        \hline
        \textcolor{black} {Step} & \textcolor{black} {Description} \\
        \hline\hline
        \textcolor{black} {Step 1} & \textcolor{black} {Select a dataset that allows for encoding of individual and classes in the binary class output matrix. This allows us to observe the effects of salience on both the individual image that is tagged with salience, as well as the class it belongs to.} \\
        \hline
        \textcolor{black} {Step 2} & \textcolor{black} {Train a SANN to classify images in the given dataset such that the network achieves 100\% classification accuracy for both the class and individual. This will serve as the \textit{baseline} model.} \\
        \hline
        \textcolor{black} {Step 3} & \textcolor{black} {Measure the impact on classification confidence for additional classification training (using back-propagation) over a defined number of additional training epochs. This will serve as the \textit{endline} model.} \\
        \hline
        \textcolor{black} {Step 4} & \textcolor{black} {Apply one-time salience training to the baseline model in various ways (as discussed above) and we compared the classification confidence and performance against the baseline and endline models.} \\
        \hline
    \end{tabular}
    \caption{\textcolor{black} {Experimental methodology}}
    \label{tab:method}
    \end{center}
    \end{table}

\newpage
\subsection{Encoder and SANN design}

\begin{figure}[ht!]
    \centering
    \includegraphics[scale=0.5]{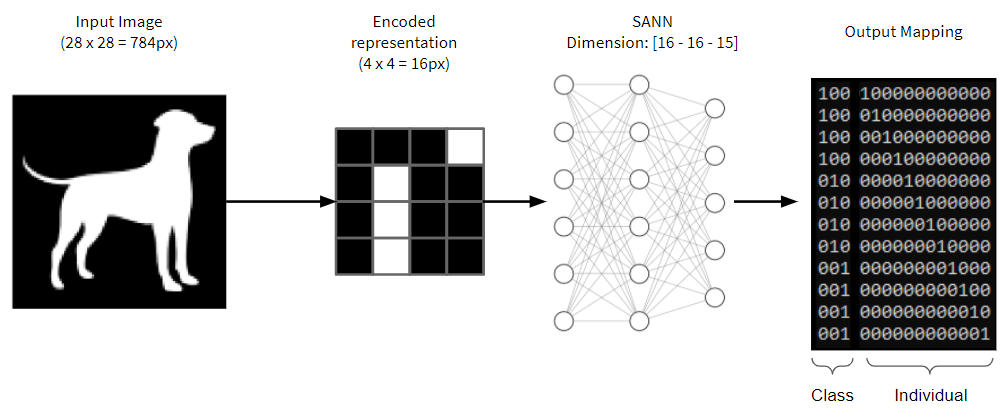}
    \caption{\textit{System architecture: Encoder and SANN design parameters}}
    \label{fig:representation}
    \end{figure}

\textcolor{black}{The design parameters of the Encoder and SANN are shown in Fig \ref{fig:representation}. Input images were scaled down to 28px $\times$ 28px and we chose an encoded representation size of 4px $\times$ 4px for computational efficiency. An encoding dimension of 16px was chosen to optimize SANN performance while still achieving an Encoder reconstruction accuracy of 93.81\% (see Section \ref{ssec:encoder_training}). We chose an output mapping of 15 binary class outputs (3 classes + 12 individuals) which we discuss in more detail in the next section.} \\

\textcolor{black}{The SANN was implemented as a fully-connected 3-layer neural network (ANN). Architectures such as LeNet architecture \cite{lecun1998gradient}, AlexNet \cite{krizhevsky2012imagenet}, VGGNet \cite{simonyan2014very}, GoogLeNet \cite{szegedy2015very} and ResNet \cite{he2016deep} \cite{he2016identity} were omitted from this study because they utilize a convolutional layer, which is out of scope of this research. The 3-layered SANN had dimensions of 16-16-15 nodes because the input was 16px and the output was 15px.} \\

\textcolor{black}{We chose to use Sigmoid classifier as the output layer because we are using binary class output encoding. While both Sigmoid and Softmax classifiers give output in [0,1] range, the difference is that the Sigmoid classifier returns an output between 0 to 1 and the Softmax classifier ensures that the sum of outputs is 1. We suggest exploring the Softmax classifier with cross-entropy loss in future research.}

\newpage
\subsection{SANN software implementation}

\begin{figure}[hbt!]
      \centering
      \includegraphics[scale=0.6]{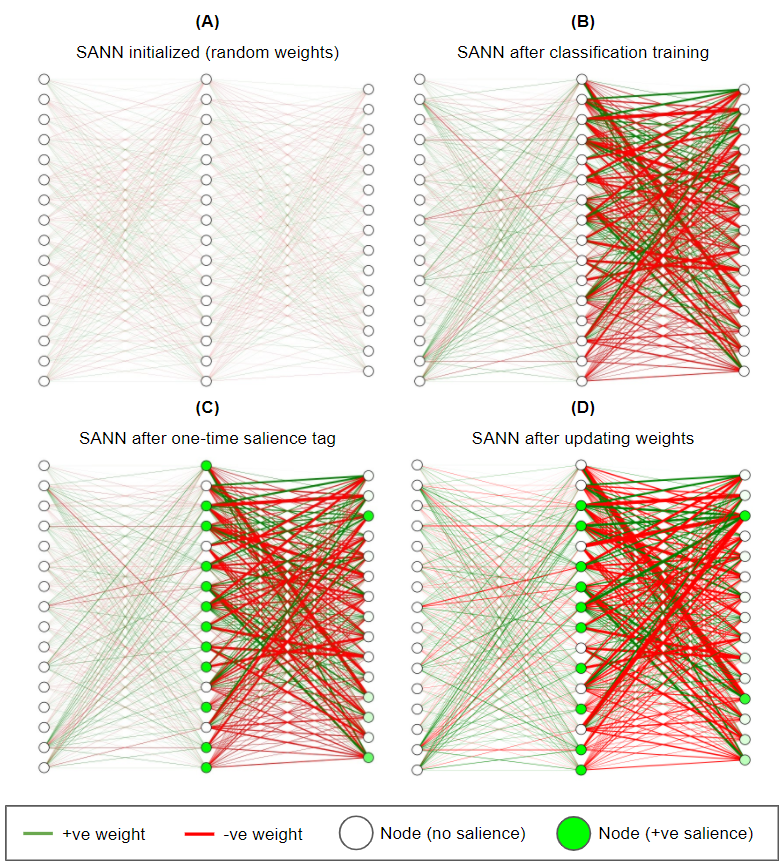}
      \caption{\textit{\textcolor{black}{A visualization of strength and salience of weights and nodes in the SANN model: (A) SANN after initialization (random weights), (B) SANN after 355 epochs of standard classification training, and (C) SANN after one-time salience training (nodes only), and (D) SANN after one-time salience update of the weights.}}}
      \label{fig:Fig006}
    \end{figure}

The ANN framework was adapted from an open-source pure python implementation of a Neural Network \cite{remmelzwaal2020python}. The activation functions of each node was sigmoidal; variations of the activation function (e.g. ELU, ReLU, Leaky ReLU) could be tested in further research. \textcolor{black}{As part of the software implementation we included a visualization tool that allows users to visualize the salience at each node in the SANN (see Fig \ref{fig:Fig006}).}

\newpage
\subsection{Dataset}

\textcolor{black}{The animal silhouette dataset was chosen to model the effects of neuromodulators released in the cortex \cite{remmelzwaal2020brainCSR}. This dataset consists of 12 x images of animal silhouettes, split across three classes: Bird, Cat, and Dog. This dataset is not as extensive as the other datasets such as MNIST \cite{yann1998mnist}, Fashion MNIST \cite{xiao2017fashion}), CIFAR10 \cite{krizhevsky2009learning} or GTSRB road sign dataset \cite{stallkamp2011gtsrb}, but is more challenging than solving a mathematical equation \cite{Thenius2013EMANN}.} \\

\begin{figure}[ht!]
    \centering
    \includegraphics[scale=0.4]{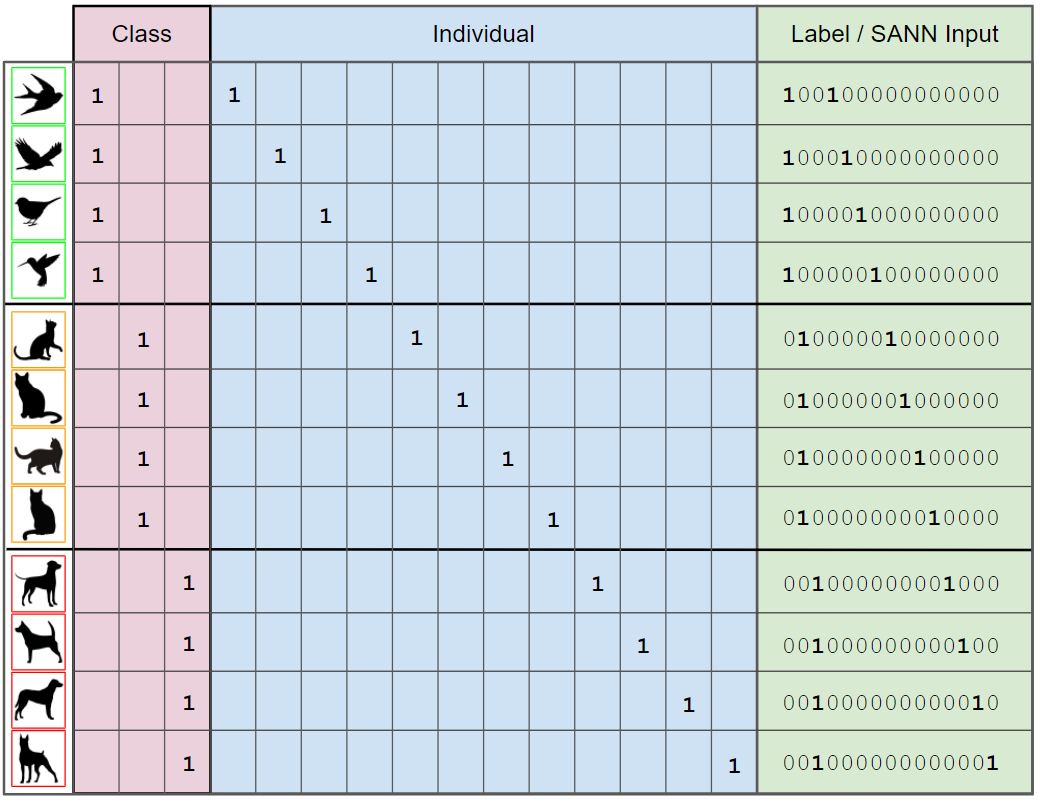}
    \caption{\textit{Mapping of image to output labels, factoring in class and individual labels}}
    \label{fig:dataset}
    \end{figure}

\textcolor{black}{The 15-element binary class output matrix was selected to capture both the class and individual mapping of images. Each input was mapped to a class element (1 of 3 options) as well as to a unique individual element (1 of 12 options). This allows us to observe the effects of salience on both the individual image that is tagged with salience, as well as the class it belongs to (see Fig \ref{fig:dataset}).}

\newpage
\section{Results} \label{sec:results} 

\textcolor{black}{In this section we \textcolor{black}{present} the results from the encoder training, SANN baseline classification training, and one-time salience training.}

\subsection{Encoder training} \label{ssec:encoder_training}

\begin{figure}[ht!]
    \centering
    \includegraphics[scale=0.45]{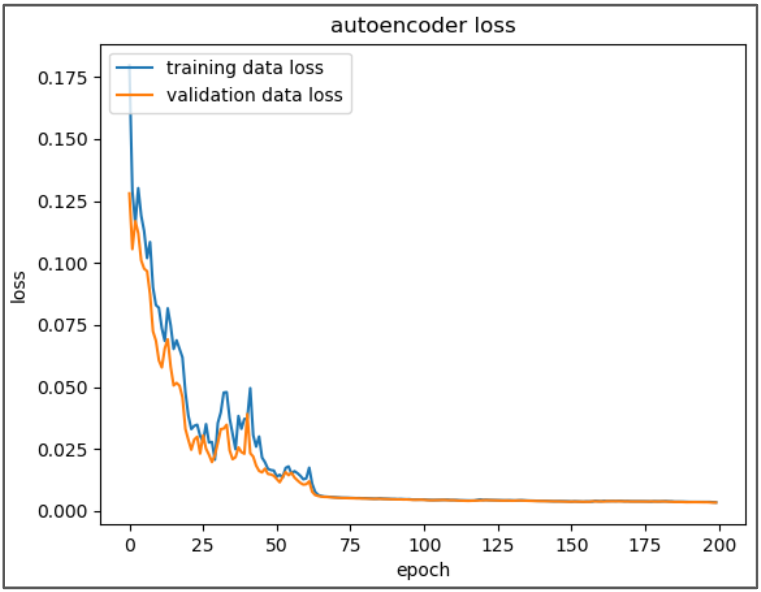}
    \caption{\textit{\textcolor{black}{Encoder loss over 200 epochs of training}}}
    \label{fig:encoder_baseline}
    \end{figure}

\textcolor{black}{The Encoder was trained with 200 epochs, and achieved an average reconstruction accuracy across the dataset of 93.81\% with a training loss of 0.0036 and a validation loss of 0.0035 (see Fig \ref{fig:encoder_baseline}).}

\subsection{SANN classification training}

\textcolor{black}{The SANN was trained on 500 epochs, and achieved 100\% classification accuracy with a validation error of 0.33695 (see Fig \ref{fig:encoder_baseline}). The SANN model achieved 100\% accuracy from epoch 355.}

    \begin{figure}[ht!]
    \centering
    \includegraphics[scale=0.35]{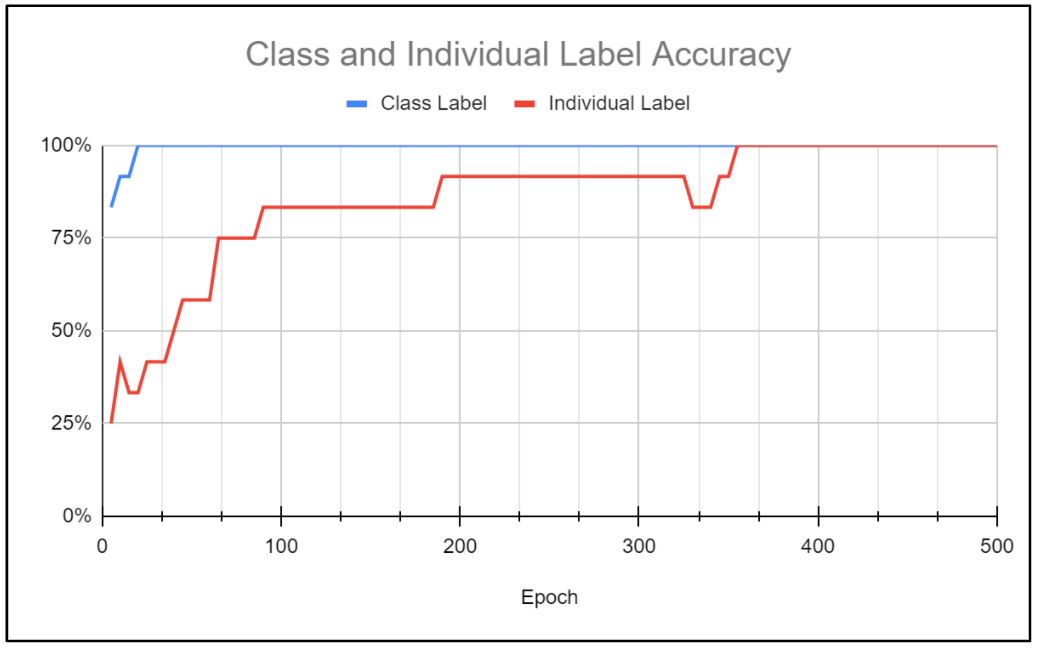}
    \caption{\textit{\textcolor{black}{SANN accuracy (for both Class and Individual labels) over 500 epochs of training}}}
    \label{fig:sann_training}
    \end{figure}

\newpage
\subsection{One-time salience training}

\begin{figure}[ht!]
    \centering
    \includegraphics[scale=0.54]{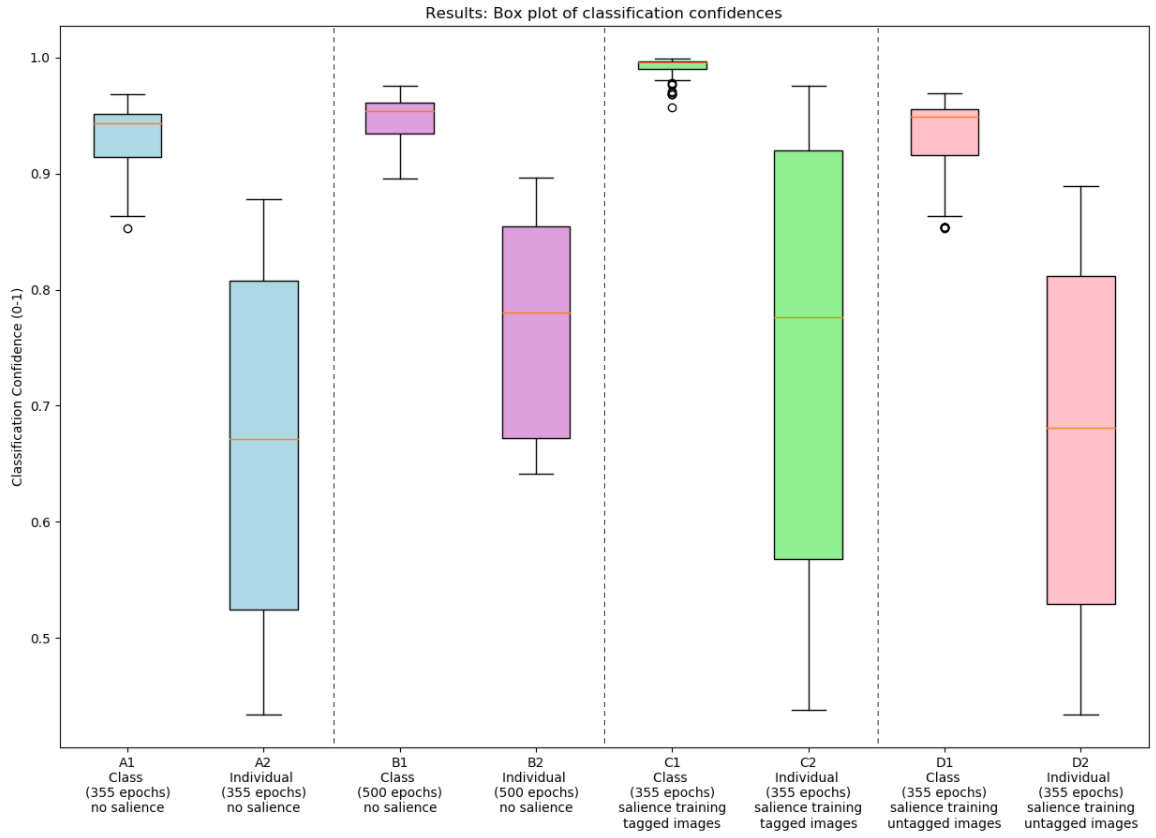}
    \caption{\textit{\textcolor{black}{Box plot of classification confidence across dataset before and after strengthening the weights proportional to the salience and node activation. This chart shows the confidences of class classification confidences (A1,B1,C1,D1) and individual classification confidences (A2,B2,C2,D2). Plots A1,A2 show confidences after 355 epochs of baseline classification training without salience and plots B1,B2 show confidences after 500 epochs of classification training without salience (benchmark). Plots C1,C2,D1,D2 show confidences after one-time salience training. Plots C1,C2 show confidences for salience-tagged images, while D1,D2 show confidences for images not tagged with salience.}}}
    \label{fig:results_boxplot}
    \end{figure}

\textcolor{black}{To model the effects of dopamine (pleasure), the SANN was subjected to one-time salience positive training after completing 355 epochs of classification training. We establish a baseline classification because neuromodulators do not create new classification patterns; the only tag existing classification patterns with salience. The results are shown in Fig \ref{fig:results_boxplot}.}

\noindent{}\textcolor{black}{From Fig \ref{fig:results_boxplot} we make the following additional observations:}

\begin{enumerate}
    \item \textcolor{black}{B1 and B2: Additional 145 epochs of classification improves the classification confidence of the SANN. This is expected behaviour of an ANN as the model begins to overfit.}
    \item \textcolor{black}{C1: One-time salience training resulted in a median confidence for the entire class that was higher than the standard SANN classification training, even after 500 epochs.}
    \item \textcolor{black}{C1: 355 epochs of baseline training followed by one-time salience training results in similar confidence of tagged images compared to 500 epochs of regular training.}
    \item \textcolor{black}{C2: 355 epochs of baseline training followed by one-time salience training results in similar confidence of images not tagged with salience compared to 500 epochs of regular training. Although there is more spread than with standard classification training.}
    \item \textcolor{black}{D1 and D2: The confidence for non-tagged images also improves after one time salience training.}
\end{enumerate}

\newpage
\subsection{Modelling the intensity of salience}

\begin{figure}[ht!]
    \centering
    \includegraphics[scale=0.55]{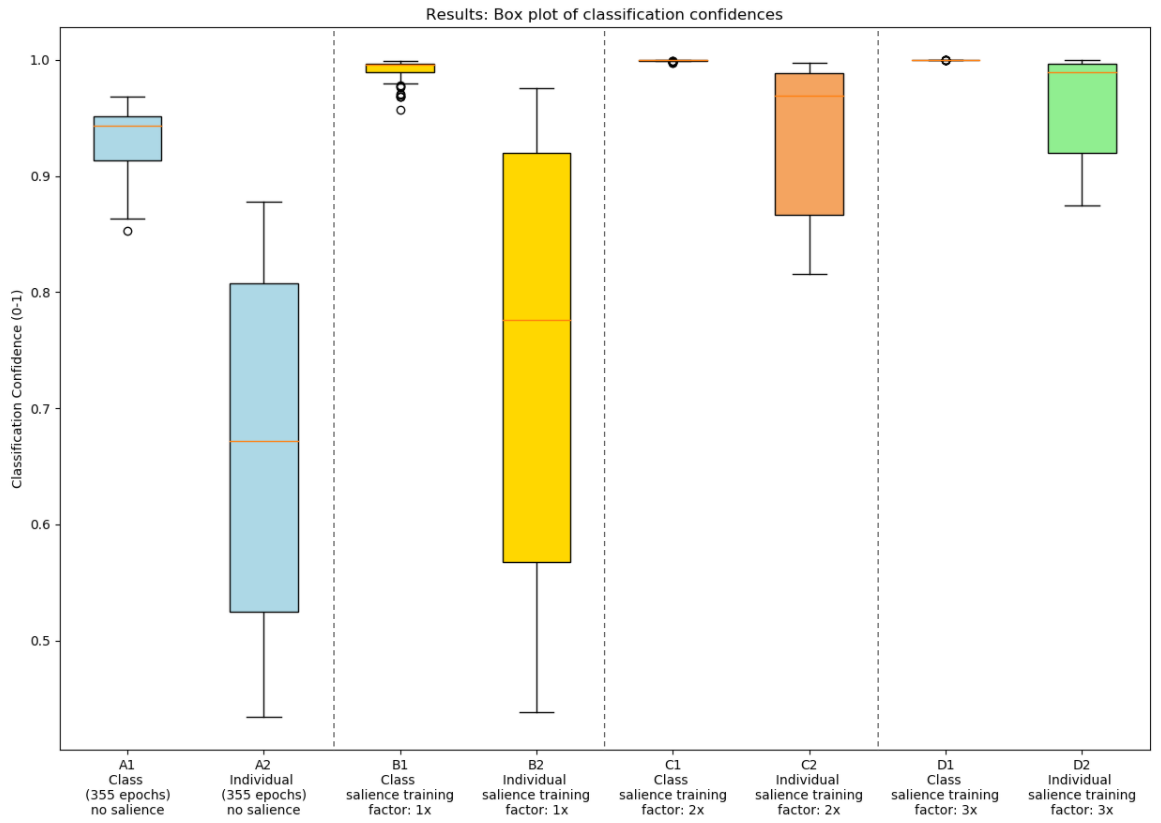}
    \caption{\textit{\textcolor{black}{Box plot of classification confidence across dataset. The SANN was tested with a salience factor of 1$\times$, 2$\times$ and 3$\times$ the baseline intensity.}}}
    \label{fig:results_boxplot_intensity}
    \end{figure}

\textcolor{black}{Neuromodulators are released in varying quantities in the cortex depending on the intensity of the experience. We model this variation in intensity by varying the intensity of the neuromodulator released during one-time salience training. To explore the effects of salience intensity, we tested the SANN with a salience factor of 1$\times$, 2$\times$ and 3$\times$ the baseline intensity. The results are shown in Fig \ref{fig:results_boxplot_intensity}.}

\textcolor{black}{The results in Fig \ref{fig:results_boxplot_intensity} show a strong positive correlation between the salience intensity and the impact one-time salience training has on the classification confidence for the entire network. However, this does call into question the effect of salience tagging on a known pattern on the SANN's ability to learn new patterns in the future, but this falls outside of the scope of this paper.}

\newpage
\subsection{Modelling negative salience (norepinephrine)}

\textcolor{black}{We modelled the effect of norepinephrine as a salience signal of similar magnitude but inverse sign. The results of this was perfectly symmetrical effect on the SANN, producing identical results, the only difference bring that the salience response $R$ was negative in sign.}

\subsection{Combining positive and negative salience}

\begin{figure}[ht!]
    \centering
    \includegraphics[scale=0.5]{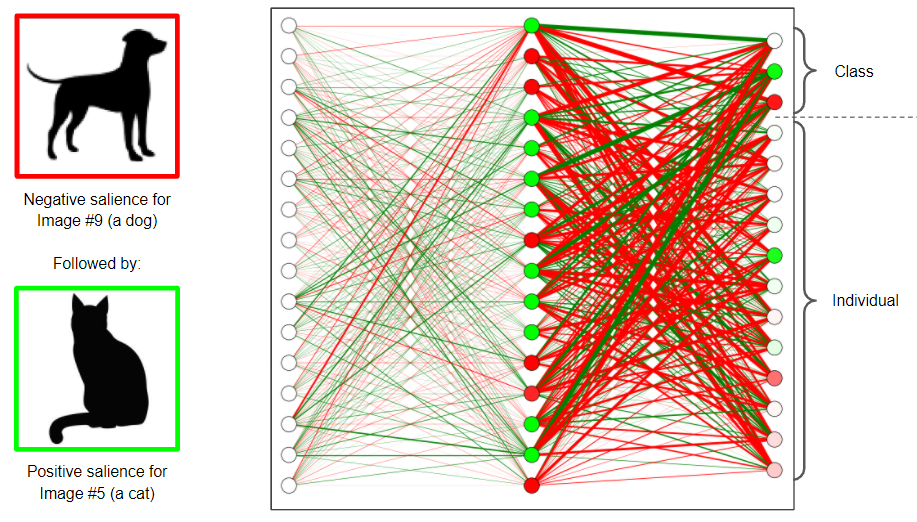}
    \caption{\textit{\textcolor{black}{A visualization of the weights in the SANN after two sequential one-time salience training events: first dopamine (positive) followed by norepinephrine (negative). The green lines represent +ve weights, red lines represent -ve weights. The thickness of the lines represent the magnitude of the weights. The green circles represent nodes with +ve salience and the red nodes -ve salience. After this training sequence we see that class 2 (cat) gives a positive salience response, while class 3 (dog) has a negative salience response. Similarly individual objects 5 and 9 have positive and negative salience responses respectively.}}}
    \label{fig:positive_negative}
    \end{figure}

\textcolor{black}{We modelled the effect of combining +ve salience (dopamine) and -ve salience (norepinephrine) as two sequential one-time salience training events. We demonstrated this on a specific pair of images, as shown in Fig \ref{fig:positive_negative}. The result showed that the SANN is capable of embedding both positive and negative salience, which is a bio-realistic observation.} \\

\textcolor{black}{We recognize that this experiment illustrated the impact of combining +ve salience (dopamine) and -ve salience (norepinephrine) for 2 separate images in 2 separate classes. We also recognize that there could be a conflict in salience signals when information is represented using overlapping populations of neurons. We suggest investigating how salience signal disambiguation could be resolved if there are overlapping population nodes as an extension of this proof-of-concept paper.}

\newpage
\subsection{Activation functions}

\begin{figure}[ht!]
    \centering
    \includegraphics[scale=0.5]{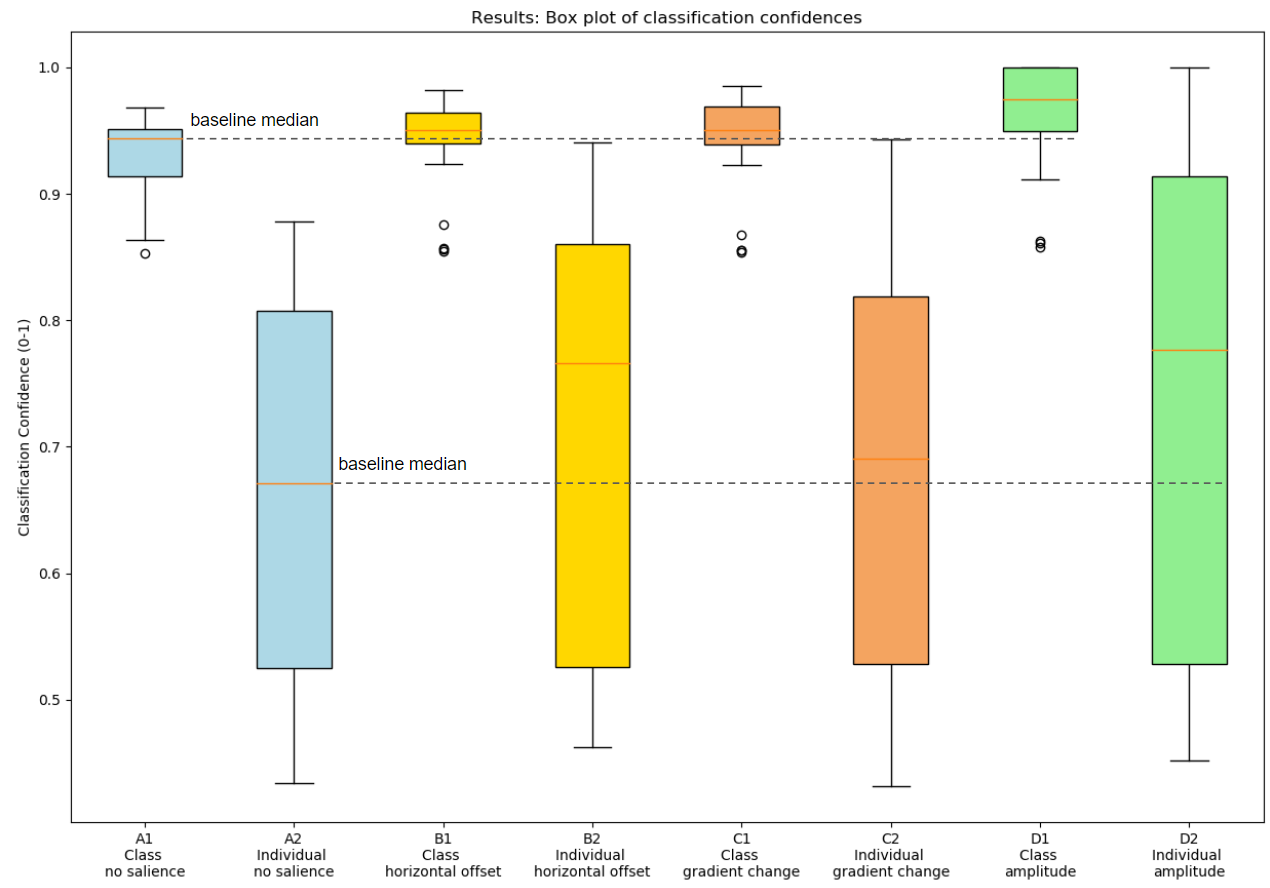}
    \caption{\textit{\textcolor{black}{\textcolor{black}{Box plot of classification confidence across dataset before and after impacting the node activation functions proportional to the salience and node activation. This chart shows the confidences of class classification confidences (A1,B1,C1,D1) and individual classification confidences (A2,B2,C2,D2). Plots A1,A2 show baseline classification confidence without salience impact. Plots B1,B2 show confidences after horizontal offset change. Plots C1,C2 show confidences after gradient change. Plots D1,D2 shows confidences after amplitude change.}}}}
    \label{fig:activations}
    \end{figure}

\textcolor{black}{After exploring the impact of salience on strengthening weights, we next explored the impact of salience on activation function. We updated the activation functions of each node only once during one-time salience training, as described mathematically in Section \ref{sec:modelling_salience}. We explored 3 variations, namely (1) horizontal offset, (2) gradient change, and (3) amplitude change. The results are shown in Fig \ref{fig:activations}.} \\

\textcolor{black}{The results show that allowing the salience to impact the activation functions of every node in the network proportional to their activation results in an improvement in the classification confidence across the entire network. The most significant improvements were seen with amplitude and Horizontal offset variations.}

\newpage
\subsection{Performance impact of salience response}

\begin{figure}[ht!]
    \centering
    \includegraphics[scale=0.4]{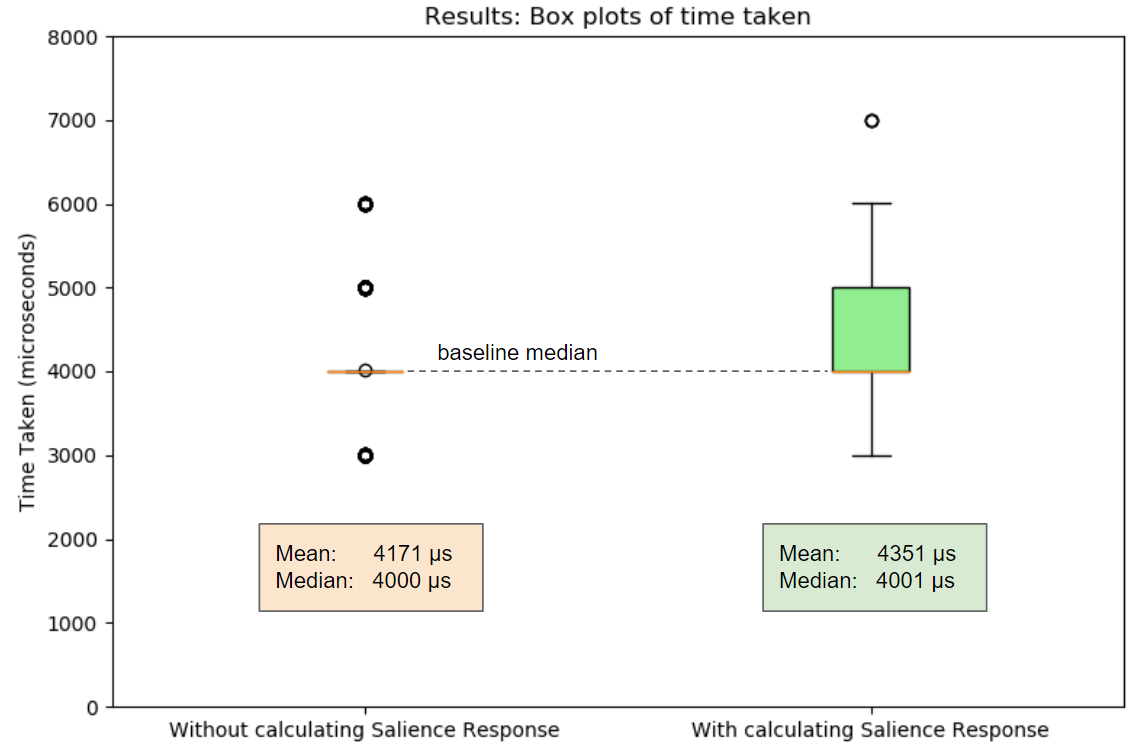}
    \caption{\textit{\textcolor{black}{\textcolor{black}{A box plot of the average time taken to perform 1200 classifications with the SANN model, both with and without calculating salience response. The median time taken was 4000µs (without) and 4001µs (with), and the mean time was 4171µs (without) and 4351µs (with). We notice an increase in the spread of calculation time with a salience response, however the mean value only increases by only 4.3\% (180µs).}}}}
    \label{fig:performance}
    \end{figure}

\textcolor{black}{Finally, we benchmarked the performance impact of calculating the salience response at the time of classification (inference). We measured the time taken for 1200 classifications, with and without calculating the salience response. The results from this test show that the median time taken was 4000µs (without) and 4001µs (with), and the mean time was 4171µs (without) and 4351µs (with). While we notice an increase in the spread of calculation time with a salience response, the mean value increases by only 4.3\% (180µs) which is a relatively small impact. The results are shown in Fig \ref{fig:performance}.}

\newpage
\section{Discussion} \label{sec:discussion}

In this paper we have introduces a new kind of Artificial Neural Network architecture, namely a \textit{Salience Affected Artificial Neural Network (SANN)}. The SANN accepts a global salience signal during training, which affects the specific pattern of activated nodes during a single iteration of salience training. This model architecture is inspired by the effects of diffuse projections of neuromodulators in the cortex. \\

\noindent{}\textcolor{black}{At a high level we recognize four key characteristics of the SANN architecture. Firstly, salience plays a crucial role in one-time strengthening of existing classification patterns in the cortex. Secondly, the SANN model successfully impacts an entire pattern of neurons simultaneously with salience when a neuromodulator is released. Thirdly, salience affects neurons proportional to their activation at the time of neuromodulator release. Lastly, a salience tag can be retrieved along with the memory, which we call a \textit{salience response}}. \\

\noindent{}\textcolor{black}{Practically, we implemented the SANN in software and made five key observations of one-time salience training in a neural network. Firstly, salience tagging has a positive impact on classification confidence across the \textit{entire} SANN after only one-time salience training. The images that were tagged with salience saw the largest improvement to both their class and individual classification confidence. Thirdly, the improvement seen at the class level was significant, meaning that classification confidence was improved for all members of the same class. Fourthly, we saw that an SANN is capable of both positive and negative salience tagging, separately and combined. And finally, the impact on performance for calculating the salience response at the time of classification (inference) was minimal.} \\

\noindent{}\textcolor{black}{The fact that one-time salience training improves the performance (in this case classification confidence) of a neural network is biologically similar. The release of neuromodulators in the cortex strengthens existing patterns and tags them with salience so that these patterns stand out from others. The fact that we see an associated salience response during inference as well as an improvement in classification confidence is what we would expect.}

\noindent{}\textcolor{black}{To highlight the significance of these findings we provide an example:} Say we train a neural network to classify images into 3 types of animal classes: cats, dogs, and birds. Furthermore, some of the image are of your favourite cat Cleo. Once the network has been trained to correctly classify images into these 3 classes of animals, you associate a positive salience with your favourite cat Cleo \textcolor{black}{and a negative salience with your neighbours dangerous dog Spike}. Associating salience this way in a SANN has 3 distinct impacts: (1) Firstly, the specific memory of your cat Cleo has a \textcolor{black}{heightened positive salience response, and the dog Spike has a heightened negative salience response}. (2) Secondly, the entire class of cats also receives a heightened \textcolor{black}{positive salience response, while the entire class of dogs has a heightened negative salience response}. In other words, associating a positive salience with your cat Cleo has a positive salience impact on all memories of your cat Cleo as well as all other cats. \textcolor{black}{(3) Thirdly, the classification confidence is improved by salience, meaning that you will be able to recognize Cleo and Spike higher confidence after salience training.} \\

\noindent{}\textcolor{black}{The impact of one-tine salience training strengthens existing pathways and connections. What is happening with one-time salience tagging is that a population of neurons that distinctly code for a high-level object are strengthened; this is analogous to a top-down influence strengthening bottom-up representations of all parts of the whole. Depending on how invariant the SANN's representations are, the same effect could propagate to images that are similar to the salience-tagged image. Next time the same image is presented, the object-relevant signal is boosted through stronger bottom-up representations. One-time salience training has a significant impact on learned behaviour, and will likely have an negative impact on generalization, and on the effectiveness of future training (e.g. fine-tuning, or transfer learning). We suggest exploring how to regularize the salience training in order to prevent memorization, which could in turn cause worse generalization. This falls outside of the scope of the proof-of-concept model presented in this paper, so we suggest it as a topic for future research.} \\

\noindent{}This paper is limited to demonstrating a proof of concept only; there are many further areas of suggested research, as well as many possible applications of this research, which we will touch on in the next section.


\newpage
\section{Future work} \label{sec:future_work}

We recommend that in future work the timing of salience training is explored, asking: \textcolor{black}{what effect would there be if salience training took place during classification training? In this paper we only observe the effects of one-time salience tagging after classification training. This research could also be extended in the future to datasets such as Fashion MNIST \cite{xiao2017fashion}), CIFAR10 \cite{krizhevsky2009learning} GTSRB road signs \cite{stallkamp2011gtsrb}, COCO \cite{Lin2014COCO}, NIST19 \cite{grother1995nist}, Chinese handwriting dataset \cite{liu2010chinese}, and facial datasets such as SCFace \cite{grgic2011face}. We also suggest exploring the impact of salience on activation functions other than sigmoidal  (e.g. ELU, ReLU, Leaky ReLU) to get a better sense of how salience signals can impact nonlinearities and their respective gradients. We also suggest exploring Softmax classifier with cross-entropy loss instead of the Sigmoid classifier. An interesting topic of future research could be exploring the impact of one-time salience training on learning generalization and on the impact of additional training (e.g. fine-tuning, or transfer learning). Extending this research to other deep neural networks (e.g. recurrent neural networks, convolutional neural networks) is also suggested as future research. We also suggest exploring how the SANN model could assist in solving the nearest neighbor problem \cite{Andoni2006near} or enhancing the locality-sensitive hash function \cite{Dasgupta2017hash}. We also suggest that the ``desire to act" value could connect with the fight or flight response in a popular cognitive architecture such as NEUCOGAR \cite{vallverdu2016neucogar}. In addition to affecting a classification neural network, this research can be extended to model the changes in synaptic strength related to long-term potentiation (LTP) which is the primary cellular model of memory in mammals \cite{Frey1997ltp}, distributed associative storage, plasticity-related proteins (PRP), or the capture of these proteins by tagged synapses \cite{Redondo2011taggingcapture}. We also suggest that the SANN is re-implemented in a standard NN framework (e.g. Tensorflow). We also suggest investigating how salience signal disambiguation could be resolved if there are overlapping population nodes when combining positive and negative salience sequentially in a SANN. Lastly,} we suggest extending the SANN architecture to explore the impact of emotion on auditory signals \cite{wang2018emotion}, attention, active inference, curiosity and insight \cite{friston2017curiosity}. What we propose could be an important part of the vision of homeostasis and soft robotics proposed by Man and Damasio \cite{damasio_machines}. In this paper we focus only on the subject of a scene, but extending this research to contextual cues would serve as a valuable extension of this research; contextual cues can disambiguate salience signals and hence develop holistic scene understanding. These suggestions all fall outside of the scope of work of this initial paper, as this paper is limited to demonstrating a proof of concept only.


\section{Supporting material}

The source code as well as records of the tests conducted in this paper are publicly available online \cite{remmelzwaal2020sann}. For additional information, please contact the corresponding author.


\section{Acknowledgements}

\noindent{A preliminary version of this work was the subject of an MSc thesis of Remmelzwaal, supervised by Tapson and Ellis. A pre-print was released in 2010 \cite{remmelzwaal2010integration}. The present version is so improved and updated that it is essentially a new paper, in particular because, apart from a greatly improved presentation of the logic of the project and its relation to brain structure and function, it has added the dynamics of a shift of weight in proportion to the salience signal and two further effects on the activation function, as well as extensive testing of how this works out in practice.}\\

\vspace{0.1in}
\noindent{We are grateful to Mark Solms, Amit Mishra and Jonathan Shock for very helpful comments, and Bruce Bassett for a useful remark.}\\

\vspace{0.1in}
\noindent{This research did not receive any specific grant from funding agencies in the public, commercial, or not-for-profit sectors.} We declare no conflicting interests.



\end{document}